# Quick Summary

By Robert Wahlstedt

Abstract

Quick Summary is an innovate implementation of an automatic document summarizer that inputs a document in the English language and evaluates each sentence. The scanner or evaluator determines criteria based on its grammatical structure and place in the paragraph. The program then asks the user to specify the number of sentences the person wishes to highlight. For example should the user ask to have three of the most important sentences, it would highlight the first and most important sentence in green. Commonly this is the sentence containing the conclusion. Then Quick Summary finds the second most important sentence usually called a satellite and highlights it in yellow. This is usually the topic sentence. Then the program finds the third most important sentence and highlights it in red. The implementations of this technology are useful in a society of information overload when a person typically receives 42 emails a day (Microsoft). Another implication is meeting summary information from video for peace officer records and sports broadcasters. The paper also is a candid look at difficulty that machine learning has in textural translating. However, it speaks on how to overcome the obstacles that historically prevented progress. This research is different from other research attempts because it takes into account heuristics in the paragraph and treats the document instead of having a list of disjointed sentences it also each sentence contributing meaning to other sentences until it achieves a pivotal point in document. Previous methods of a document summary generator included reducing redundant words or junction words until it develops a nuclear core. This paper proposes mathematical meta-data criteria that justify the place of importance of a sentence. Just as tools for the study of relational symmetry in bio-informatics, this tool seeks to classify words with greater clarity. "Survey Finds Workers Average Only Three Productive Days per Week." Microsoft News Center. Microsoft. Web. 31 Mar. 2012.

Introduction

A piece of literature is similar to a piece of music. While the BBC has a recording of Tchaikovsky's 1812 overture that lasts 17 minutes, most people remember only the theme. In this blog, we discuss what makes literature themes stand out from the rest of the piece through repetition, supporting variations of the theme. Many textbooks of text mining explain the author's beliefs that word categories or parts of speech are important. These papers find summaries with the hope that by deleting irrelevant words until only the nuclear core is remaining. However, these authors overlook the importance of context. It is important to use latent semantic analysis. This seeks to use structural knowledge of pragmatic. For example, to get to the best diagnosis of a patient, it is important to take into account their medical histories and not just a few quotes. This method does analysis on the entire document before concluding so metaphors and words with multiple meanings are not confused. This paper proposes that by using meta-data, or data about the data, achieves the goal of distinguishing repetition of the theme and supporting variations also known as satellites. This is achievable by finding the roots of words through their origins, translating the word prefix, root, and suffix into meta-data.

Designing with the Brain in Mind

In designing a program, it is necessary to understand what one has to work with. Imagine a machine with two cameras perceiving words. When a person reads a word like "unbuttoning" the human brain strips off the "un" and the "ing" and it is left with two morphemes, "but" and "ton". There is a priming effect where a person is more inclined to think about a certain flow of words. For example, after casa, there sometimes is the word Blanca like the movie.  This is why we understand differing homograph meanings. A grapheme is a letter or a series of letters that map a phoneme in the target language. Imagine a web of words that links all words in existence.  In the book, The Neuroscience of Language introduces a word-related functional web. It is a synchronous firing chain or synfire chain. At first glance, automata might be similar to this web. However, a computer differs from a brain because even the multi-core processors are uni-core in the sense that the computer can only process a few threads unified by a central process.  The human brain is a democracy having processes happening simultaneously and the strongest impulse, usually by a collective number of neurons fires that is what the human brain thinks.  Miguel Nicolelis argues in his book, *Beyond Boundaries*, there is not a separate part of the brain language support.  However, a certain sequence of neurons that light up when a person reads about an object because it invokes an emotion.  The book, The First Word, describes how non-humans animals could use simple language.  Instances of non-humans are territorial warning signs and primitive inner-species communication.  The article "How Dolphins Say Hello" says that dolphins only whistle when they are in a group of dolphins and identify themselves through echolocation.

Grammar is a set of rules in a language that allows people to communicate and understand.  There is a debate between B.F. Skinner and Noam Chomsky. Skinner believes that a person learns language through association, the sight of things along with the sound of the word.  This association is reinforced to match conform to a dialect.  Chomsky believes that a Martian scientist observing children in a single-language community would observe that language is very similar regardless of the culture and therefore innate.  Although reading is unnatural, our brain wants to see patterns and group objects together. We recognize lines and shapes that we can recognize characters. We then group these characters into morphemes. We then put these morphemes into phrases, idioms, words, and sentences. These sentences form a paragraph. Similar to computers recognition is easy, recalling meaning is difficult.

Similar Research

Text mining projects at Universitat Politecnica de Catalunya depend on lexical categories for their lexical analysis tagging. In the book, *When you catch and Adjective, Kill It* suggests, "According to grammarians, adjectives, nouns, verbs, and injunctions are "open" parts of speech because they shift functions … and because new words are continually added to their ranks. This proves that classification of words is not always necessary." (Yagoda, 43). To make matters worse, a gerund is using a verb in its –ing form as a noun such as living. Although the English language has synoptically, there are two types of words, content words and structure words. Structure words are eliminated such as "hope that" "clearly" "strangely" "indeed" "conceivably" "seriously" "ultimately" "theoretically" " naturally"" ironically"" fortunately"" incidentally" do not containing meaning in themselves however they send signals to listeners about words that are coming later. The word "like" is overused. The National Science Foundation did a study that demonstrated when a storyteller used the word "this" instead of "a" or "an" the person had better retention. Another concern is that these systems do not account for pragmatic that is the study of context is it situational.

English as A Moving Target
Another criticism is that English is a moving target. English is a West Germanic language that borrows its alphabet from Latin, adding three characters J, U, and W. During the Norman Conquest of England the French brought over a morphed version of Latin that morphed into English. Languages morph over time and words take on other meanings. For example, consider the common phrase Santa Claus. It began as "Sant Herr Hiclaes" in Dutch, transforming to "Santerclaes", and eventually became the English "Santa Clause" today (McWhorter *the Power of Babel*, page 29). John Dryden admitted to translating his works to Latin to get the syntax to flow smoother (Lynch, 36). The two types of people who study are either prescriptive grammar or descriptive grammar. Scholars describe Samuel Johnson, the writer of

A Dictionary of the English Language, as a descriptive grammar lexicographer. He realized there could be no establishment that could enforce grammar. Printers noted that they could sell to more people their books if there was a standard way of speaking. Later, the success of The Oxford English dictionary benefited from a correspondence between the editors and the public. Today is no different. French President Sarkozy received a concerned note saying that the recipient was worried that French was on a downward spiral linguistically and Sarkozy should do something (Greene). China had the "Law of the People's Republic of China on the Standard Spoken and Written Chinese" that requires media personal and broadcasters a certain level of speaking ability. In December 24 2011, this ban lifted. Linguists believe that all languages are connected. Instead of being different "languages", they are dialects of one common language that originated from the Persian Gulf.

Ambiguity of what is a Summary

A recent survey done at the University of Illinois showed that if surveyors present a passage of text to a human subject they are likely to disagree on what passage are a summary. This is because culture makes the subject biased and the subjects project their preconceived notions on the passage. This shows the benefits of an artificial intelligence system that notices pronouns that a person who overcome in emotions fails to recognize. In a study by James Pennebaker at the University of Texas Austin, he describes how natural language processors can detect who might win an election. In his book, *The Hidden Life of Pronouns* he accounts of how aides told John Kerry to use the pronoun "I" less. Should a person be of the same political party as John Kerry, they might overlook the fact that he was using words in this format. Linguists call this connotation lexical semantics, the study of what words denote.

How to Address Critics who question the standard of English spelling

The Language Wars by Henry Hitchings discusses how it is amazing that people spell words uniformly. Before computers, he points out that even dictionaries were inconsistent. There was a survey where a computer used phonetics to spell out words and 50 percent of the words did not agree with the phonetic spelling. Steven Pickner tells us that 84 percent of the words have spellings that conform to patterns we can notice. As shown by these statistics, the ability to conform is possible.

Proposal for Meta-data

It is necessary to study of meta-data to correlate etymology with words. Etymology is the study of finding word roots and morphology is the study of word making. Today advances in bio-informatics tools have yielded a complex study of the human genome and have made great strides in finding out mutations that can create a higher chance of cancer. We should do so because we can learn about the human race as well as expand our knowledge of words. Here are some examples: nickname originally came from nekename containing the compound eke and name. Eke came from ēac that means also. It means an additional name. Hobby comes from the word hob and yn. Hob or hōb means a threaded and fluted hardened steel cutter, resembling a tap, used in a lathe for forming the teeth of screw chasers, worm wheels (Webster's Revised Unabridged Dictionary) and yn means "to be". It means to spin time. Omelet comes from the French word omelette. Omelette comes from alemette that comes from alemelle that comes from la lemelle that means a thin plate like structure. We should know how to find the prefix, root, and suffix because we are able to understand why someone in history responded to situations. The written word often transcends a person's lifespan and the culture changes. A written manuscript is similar to a fossil. For example, rabbit comes from the words robète that means to steal. This explains why Mr. McGregor thinks Peter Rabbit as such a nuisance. If you are a person that is a Christian, it might please you to know that Nicodemus comes from the word nincompoop. This comes from the Latin phase non-compos mentis meaning not of sound mind. People regard him as a hero of the Christian faith because he walked by faith and not by knowledge.

What does a computer have to do with this? Computers are able to carry out arithmetic or logical operations such as regular expressions or pattern matching. Given a rules file and a lexicon file, they are able to study morphology.

Examples of Meta-data

"A" as in rag or Prague is big and "i" as in pin is small, "e" is somewhere in between. There has been some big vowel shifting in England between 1350 and 1500 known as the big vowel change so words like big and small have the wrong letters. Little and large are right. Pimpf means "a little boy", pimple means a little swelling, and pampers means a lot. Servant is in between. Quick Summary uses Maven Meta-data Markup Language or MMML. It has a structure similar to the one shown below.

```xml
<?xml version="1.0"?>
<wordlist>
  <word id="daisy">
    <word>daisy</word>
    <origin>Latin</origin>
    <source>dæges ēaġe</source>
    <morphemes></morphemes>
    <sentencelastused> I picked for you a daisy.</sentencelastused>
  </word>
</wordlist>
```

Artificial Neural Networks

As discussed before in this section designing with the brain in mind, automata is not an adequate algorithm for defining morpheme parsing. A neural network a statistics algorithm used in machine learning. It is simple yet effective. Biological neural networks inspire artificial neural networks. They are adaptable and self-organizing for real-time data. They also give fault tolerance. A neural network is based on a model that works. Consider the images below.

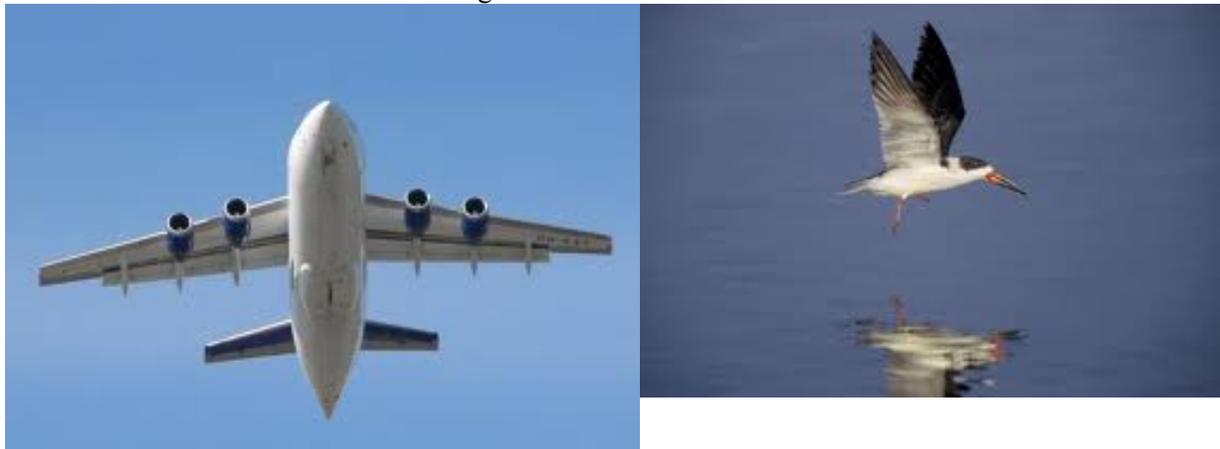

These are both images of a flying object. The one on the left is inspired by the one on the right. They both have two wings and a tail and are both of the same air-stream figure. When the inventors like Leonardo De Vinci was constructing a flying machine it very much resembled a bird. Similarly have a natural model of a brain (Abu-Mostafa). It is self-organizing, fault tolerant, and adaptable. We need to build a software design that resembles this working model. Looking at the PyBrain code it looks as though we can and should adapt something like this for our sentence processing sequence. In *How the Brain Works*, Steven Pinker discusses the benefits of using a neural network although he says the logic could be improved for multiple instances of the same object.

Language based Game Theory

A project in the Netherlands led by Jan Dietz seeks to map words as they relate to events in the world. Their premise is that we are similar to parrots repeating ideas that come to us through movies and books. In a game of broken telephone, a group of people sits in a circle and whisper a message to the person next to them. The message tends to vary as the first person conveys the message to the next person. These researchers claim through recursion we repeat these events demonstrated to us in a particular order. They believe they can forecast behaviors based on the past. The DEMO research group claims that BPMN, UML-AD, EPC, IDEF-3 are method independent. They do not take into account the redundant nature of business. It captures the conversations between actors. Jan Dietz explains that the subject chose to share one of their thoughts called the locutor. It appears as

<locutor> : <illocution> : <addressee> : <fact> : <time>

For example, let us consider checking a book out from a library. Certain circumstances must satisfy the checkout conditions. First, the subject must be a member at the library. Then he must not have any pending fees. Essence, the attributes that make the subject what he is. He is a candidate who can pick books out of the library. Should there be invalidation, he cannot do the transaction that is to pick up books. The production act of picking books out of the library is the executioner. It is a repeatable act. David Bellos tells an account of how his father was able to check into a hotel in a non-English speaking community because both he and the hotel staff were familiar with the process. In the Story of English in 100 Words, David Crystal discusses the origin of words. He gives the history of the word jail borrows from the French originating about the time of the Norman invasion of 1066. This we can tell because the word gaol is Latin. Crystal explains that this word was borrowed twice or double borrowed. We can tell this because the word did not get its original meaning. This is also true for the word convey originality coming from the French word convoy. Our approach is to take this research and combine it with what is known about UML, I* and the Zachman framework to discuss the questions of what, where, when, why, who and how. With this data, it is possible to verify that our summary correctly reflects the emphasis of the paper.

In *Empires of the World: A Language History of the Word*, Nicholas Ostler demonstrates beginning with the ancient Babylonians carrying off the Hebrew slaves and forcing them to melt into the culture of the Babylonians how language can melt together. John McWhorter writes this is what happened during the Norman invasion. There was not a bloody war instead slowly, over time the Normans introduced French. English, Crystal says, is like a vacuum, sucking up words around it. McWhorter likens it to summer camp, where an exchange of ideas takes place. In *Globish*, Robert McCrum writes that non-native speakers to surpass the limitations of English by borrowing from other languages use globish. This field of anthropology is trans-cultural transfusion. This is an example of diffusion by choice.

By studying hyperdiffusionism under the belief that man originated in one place, we can learn through cultural similarities. For instance, most civilized cultures today teach us that it is wrong to kill or commit adultery. Noam Chomsky writes about a universal grammar being indistinguishable by a Martian linguist. McWhorter claims that through DNA evidence anthropologists discovered the village where the Germanic dialect began. In tracing word etymologies, it is important to ask ourselves where features of a language began. Guy Deutscher makes the following observation: "Often, it is only the estrangement of foreign tongues, which their many exotic and outlandish features, that brings home the wonder of language's design." By incorporating origin into a NLP translator, we can see by parapraxis, images that the speaker has in mind. Through a more careful study of word choice, we see the advantage of seeing the speaker's words and come closer to making up for the disadvantage of not seeing him communicate those words to us. We can see how a language would transform and predict new words that might occur.

This study is neurolinguistics.

Questions for Further Research

Although this project covers many bases, it is important to remind ourselves of what a sentence is. A sentence is as a flatbed truck here is nothing of value until different pieces by the author to the pile. These pieces consist of verbs, nouns, and conjunctions that are all linked together.  There are different types of verbs including sensory verbs that express information an author receives from their senses. There are action verbs that describe an action, such as, can you … for me. There are infinitive that such as "to be".  These words just hold the sentence together. A sentence is made out of declarative that is a declaration that make statements. Interrogative is a question and are sometimes rhetorical.  Imperative is a request or demand some action such as a how-to.  Exclamations express strong emotions. Sentences can be either negative or positive based on logic in the sentence. Sentences are either active or passive. An active one is happening now, and the passive occurred some time ago.  Bearing this in mind, we can question what to do about coordinate conjunctions.  This is separate from having a comma in there because the phrases on the two sides of a semicolon are able to stand independently and complement each other often to contrast things. However, like a company, one is like a boss and the other one is a subordinate conjunction or subordinators. A noun class consists of a subject or predicate normative. There might also be a relative adjective clause.  Sentences with semicolons are sometimes called swinging gates. In addition, there should be consideration to error correction of the paragraph.  Can the program shave off propositional phrases? How can the program spot and shave off introduction words that do not add meaning of a sentence and are before a comma.  What about paragraphs that are not well-formed and have preposition stranding or dangling infinitives or a dangling modifier? What if there are intentional fragments? How should the program treat sentences missing elements such as "a" or "the"? How should the program approach paragraphs that lack unification? These do not revolve around one topic. How should the program handle meandering sentences?  These could be long, rambling sentences for a powerful sensory experience.  This is an intentional run-on-sentence. Another warning is to have concise text but not so concise that it lacks the author's voice nor sounding mechanical and alienating.

Conclusion

English text is different from English speech because text is unnatural. It is important to think about approaching text as speaking onto the page and less of a production. Writing is a chance to step back from the writing and become detached, driven by ideas. By simulating what is natural and functional including neural networks speaking and listening people are able to devise methods of thinking about how and why people generate speech and analyze the speaker's word choice.